\documentclass[conference]{IEEEtran}
\IEEEoverridecommandlockouts
\usepackage[colorlinks,urlcolor=blue,linkcolor=blue,citecolor=blue]{hyperref}
\usepackage{color,array}
\usepackage{graphicx}

\usepackage{mathtools}
\usepackage{algorithm}
\usepackage{algorithmic}
\makeatother
\usepackage[utf8]{inputenc}
\usepackage[table]{xcolor}
\usepackage{tabularx,booktabs}

\newcolumntype{C}{>{\centering\arraybackslash}X} 
\usepackage{cite}

\usepackage{amsmath}
\usepackage{amsmath,amssymb,amsfonts}
\usepackage{algorithmic}
\usepackage{graphicx}
\usepackage{textcomp}
\usepackage{xcolor}
\usepackage{hyperref}
\usepackage{caption}
\usepackage{pifont}
\usepackage{xcolor}
\usepackage{rotating}
\usepackage{rotfloat}
\usepackage{algorithm}
\usepackage{algorithmic}
\usepackage{gensymb}
\usepackage{textcomp}
\usepackage{placeins}
\usepackage{balance}
\usepackage{subfigure}
\usepackage{booktabs}
\usepackage{tabularx}
\usepackage{array}
\usepackage{pbox}
\usepackage{longtable}
\usepackage{booktabs}
\DeclareUnicodeCharacter{2212}{-}
\def\BibTeX{{\rm B\kern-.05em{\sc i\kern-.025em b}\kern-.08em
    T\kern-.1667em\lower.7ex\hbox{E}\kern-.125emX}}
\begin{document}

\title{Medical Image Segmentation using LeViT-UNet++: A Case Study on GI Tract Data. }

\author{\IEEEauthorblockN{1\textsuperscript{st} Praneeth Nemani}
\IEEEauthorblockA{\textit{Dept. of Computer Science and Engineering} \\
\textit{IIIT Naya Raipur}\\
Raipur, Chhattisgarh, India \\
praneeth19100@iiitnr.edu.in}
\and
\IEEEauthorblockN{2\textsuperscript{nd} Satyanarayana Vollala}
\IEEEauthorblockA{\textit{Dept. of Computer Science and Engineering} \\
\textit{IIIT Naya Raipur}\\
Raipur, Chhattisgarh, India \\
satya@iiitnr.edu.in}}

\maketitle
\begin{abstract}
Gastro-Intestinal Tract cancer is considered a fatal malignant condition of the organs in the GI tract. Due to its fatality, there is an urgent need for medical image segmentation techniques to segment organs to reduce the treatment time and enhance the treatment. Traditional segmentation techniques rely upon handcrafted features and are computationally expensive and inefficient. Vision Transformers have gained immense popularity in many image classification and segmentation tasks. To address this problem from a transformers' perspective, we introduced a hybrid CNN-transformer architecture to segment the different organs from an image. The proposed solution is robust, scalable, and computationally efficient, with a Dice and Jaccard coefficient of 0.79 and 0.72, respectively. The proposed solution also depicts the essence of deep learning-based automation to improve the effectiveness of the treatment.  
\end{abstract}


\section{Introduction}
Medical image segmentation \cite{9363915} can be broadly defined as extracting the regions of interest in an image, such as body organs, lesions, and tumors. The principal objective of medical image segmentation is identifying and localizing critical areas of the anatomy required for effective treatment or research \cite{9418385}. However, the ineffectiveness of manual image segmentation could be inferred from its monotonous and time-consuming nature, less accuracy, and factors like variation in modalities and increased data generation. This can be inferred from the fact that it has a wide range of applications like the study and diagnosis of different ailments, including kidney tumors, cysts, skin lesions, breast cancer, brain tumor, and Gastro-Intestinal Tract cancer. Therefore, it is essential to devise techniques to extract the desired region of interest with no manual intervention to increase the effectiveness of further diagnosing different ailments. 

\begin{figure}[ht]
    \centering
    \includegraphics[width = \linewidth]{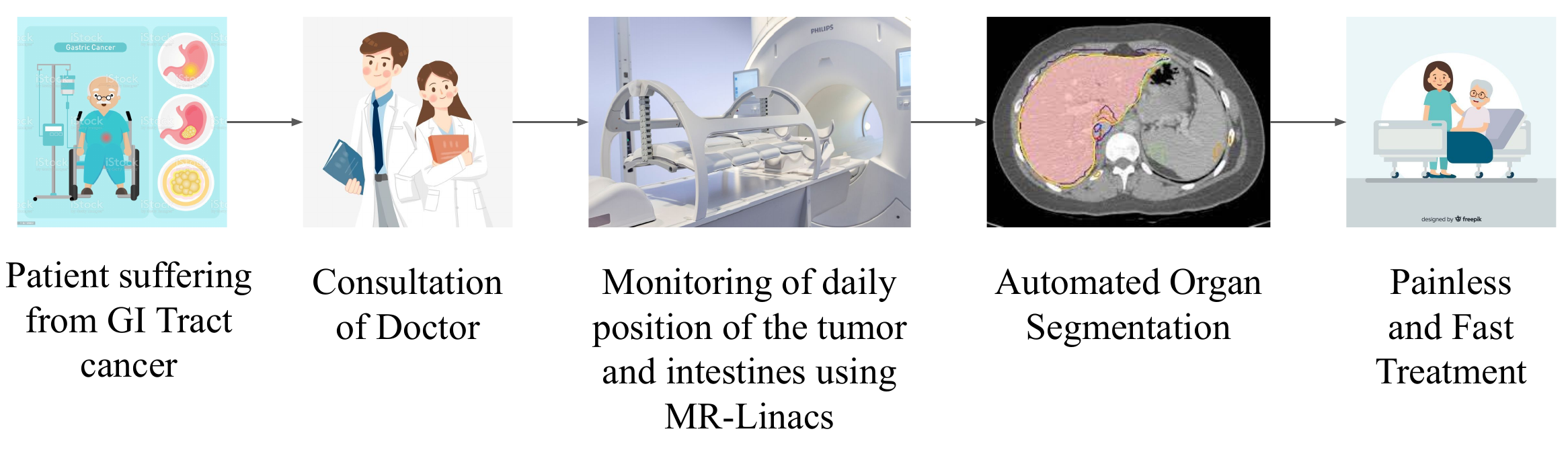}
    \caption{The Conceptual Overview of the Proposed Solution}
    \label{fig:CO}
\end{figure}

Gastro-Intestinal Tract cancer \cite{heavey2004gastrointestinal} is considered a fatal malignant condition of the organs in the GI tract, including the stomach, intestines, pancreas, rectum, and anus, with a relative survival rate of 32\%. It is estimated that 5 million people worldwide were diagnosed with this kind of malignancy in 2019. However, research has also shown that about 50\% of the diagnosed people seemed to be eligible for radiation therapy that is usually performed for 10-15 minutes per day ranging from 1-6 weeks. To treat this ailment, high portions of radiation using X-Rays are performed by radiation oncologists avoiding the regions containing the stomach and intestines. Owing to the development of new technologies like MR-Linacs, also known as magnetic resonance imaging and linear accelerator systems, day-to-day monitoring of the organs' position is possible with changing positions of the organs. However, the manual tracing of the organs' position is considered time-consuming, accounting for 15 minutes - 1 hour for every image. Hence, deep learning-based automation could form the basis for enhanced treatment with minimal time. Fig. \ref{fig:CO} shows the conceptual overview of the proposed solution. 

Image segmentation and processing form the basis of many computer vision studies \cite{9356353}. The concept of representing an image in a matrix of numbers and its division into several segments homogeneously has resulted in the development many state-of-the-art algorithms for processing and inference. Also, integrating deep learning techniques for prediction with computer vision and the high computational specifications of today's systems have resulted in the development of proficient systems yielding high-performance results for medical image segmentation. 

Convolutional Neural Networks, also known as CNNs, have formed the basis for many state-of-the-art image segmentation and classification solutions \cite{sarvamangala2021convolutional}. They are considered very effective due to their ability to reduce parameters while retaining the model's quality. Due to their densely connected architecture, they are very effective in image segmentation tasks involving high-quality images. Modern-day methodologies imply the usage of transfer learning to solve an image segmentation task. However, their inability to record the relative positions of features makes CNNs inefficient in tasks requiring an attention mechanism. 

Owing to the substantial dimensional properties of images, there is a need for receptive fields to track long-term dependencies within an image. This could be performed by increasing the size of the convolutional kernels. However, the major limitation is a reduction in the computational efficiency compared to the local convolutional operation. Hence, there is a need for a self-attention mechanism where each element of a sequence can interact with every other element present to decide upon the weightage given to each of them. So to perform this operation, we imply the usage of Vision Transformers (ViT) \cite{khan2021transformers} by combining the hard inductive bias property of CNNs \cite{d2021convit}. In this work, we intend to perform medical image segmentation by introducing LeViT \cite{graham2021levit} as the encoder of the LeViT-UNet++ architecture, thus improving the transformer block's accuracy and efficiency. The significant contributions of the proposed solution are stated as follows:

\begin{itemize}
    \item In this work, we provide a cognitive study of the techniques used for medical image segmentation on both known and unknown domains. 
    \item We also use the state of the art techniques for image preprocessing and metadata inferences.
    \item A Novel hybrid CNN-Transformer model is proposed to segment the organs from an image. 
    \item To the best of our knowledge, this work is highly scalable and computationally efficient on a wide range of domains with good performance. 
\end{itemize}

\section{Related Research Overview}
Automated medical image segmentation has always been a pioneering topic of research amidst researchers since the 19th century. One of the earliest proposed solutions includes the application of edge-preserving filters and active contour models \cite{1286732}. The subsequent solutions proposed for noise removal and effective segmentation of medical images include particle filtering \cite{6188759}, bilateral filtering \cite{4353787}, and diffusion filtering \cite{4313303}. However, the rise of deep learning and the enhanced computational capacity of present-day computers have replaced the traditional machine learning applications involving the usage of handcrafted features \cite{hesamian2019deep}. The evolution of deep learning and efficient image processing techniques have paved the way for state-of-the-art solutions for identifying the desired pixels of both organs and lesions with minimal background noise. In the following aspects of this section, the overview of the different techniques involved in the segmentation is presented.

The concept of fuzzy clustering involving the presence of a data point in more than one cluster has gained significance in many deep learning classification tasks today. Keeping this view, Devnathan et al. \cite{9505562} proposed a Fuzzy C-Means clustering technique validated on the ISIC-2018 Skin Lesion dataset \cite{9719383}. The proposed algorithm showed a significant improvement in performance as compared to the Superpixel-based Segmentation algorithm. Similarly, Beddad et al. \cite{8966821} proposed a spatial fuzzy C-means Clustering method for removing noise and heterogeneous intensity pixels. This approach was implemented with the Simulink on the brain MRI image data. 

A deep learning approach for medical image segmentation on unseen domains was proposed by Zhang et al. \cite{8995481} This work involves a deep-stacked transformation approach for domain generalization. The authors emphasized the proposal of an approach that does not require the training of the DL model repeatedly while being robust to several unseen domains. The proposed approach was validated on the ISBI LiTS 2017 liver segmentation dataset and a publically unavailable dataset for spleen segmentation. Similarly, Chen et al. \cite{8988158} proposed a solution involving a bidirectional cross-modality adaptation for unseen image domains. In this work, the authors proposed a framework known as Synergistic Image and Feature Alignment (SIFA) for unseen domain adaptation. The transformation of images is performed using generative learning with a deep supervised mechanism. The proposed solution was validated on the Brain MRI and CT data. 

Recent works on CNNs show that the model's accuracy and performance could be improved if they contain densely connected layers. In this context, Gridach et al. \cite{9434072} proposed Dopnet: a densely oriented pooling network to capture divergence in size of the features while preserving spatial interconnection based on the ideas of dense connectivity and the pooling oriented layer. Kaul et al. \cite{8759477} proposed Focusnet: an attention-based encoder architecture that can be integrated with densely connected CNNs. This approach was validated on skin lesion and lung lesion segmentation datasets. 

One of the pioneering CNN-based solutions for medical image segmentation was introduced by Ronneberger et al., also known as U-Net \cite{ronneberger2015u}. This work is considered a state-of-the-art solution due to its improved performance on datasets having a minimal number of images. The architecture of the proposed solution consists of a contracting path for the context extraction and an expanding path that facilitates precise localization. The significant advantage of this solution is its ability to segment extensive data with reduced computational time. The proposed framework was used to segment neuronal structures in electron microscopic stacks. However, the limitations of this solution included the unknown optimal depth of the architectural paths and the improper skip connections, leading to an unnecessarily restrictive fusion scheme. To address these limitations, Zhou et al. proposed U-Net++ \cite{8932614}, an enhanced version of the existing U-Net architecture that defines the unknown depth by an efficient ensemble of U-Nets with simultaneous co-learning. Also, the skip connections are redesigned so that various features are appropriately aggregated. This approach was validated on six benchmark datasets, and the results showed improved performance and computational time. 

Recently, many solutions have been proposed to improve the accuracy and efficiency of U-Net and U-Net++. Jha et al. \cite{8959021} proposed a hybrid architecture of ResUNet++ that unravels the integration problem by integrating the property of residuality with the existing U-Net++ architecture. The proposed solution validated on the Kvasir-SEG and CVC-ClinicDB datasets depicts a higher performance than U-Net and ResUNet. Similarly, Yin et al. \cite{9118916} proposed a deep guidance network based on the U-Net architecture with a guided image filter module that preserves the configuration through the guidance image. This approach achieves good results on four publically available datasets wth less training and inference time. 

Considering the efficiency of vision transformers (ViTs), several works involving a hybrid integration of ViTs and CNNs were proposed. Feng et al. \cite{9776250} introduced UTransNet: a hybrid mechanism involving the fusion of the transformers' self-attention mechanism with the existing U-Net architecture. The methodology was tested on the ATLAS datasets and has achieved a higher accuracy with fewer parameters. Huang et al. \cite{9750648} proposed TransDE: an architecture that integrates transformers with a double encoder mechanism. The dual encoder consists of global and local encoders involving transformers for sequence-to-sequence and VGG-19, respectively. The methodology was tested on the CVC-ClinicDB dataset with an accuracy improvement of 1.97\%. 

However, the majority of the above solutions require heavy computational requirements for their execution. In our solution, we employ the usage of transfer learning to make our solution computationally efficient. The overview of the existing literature is summarized in Table \ref{related}.  

\begin{table}[htbp]
\caption{Overview of the existing solutions}
\begin{center}
\resizebox{\columnwidth}{!}{%
\begin{tabular}{|c|p{42mm}|p{32mm}|c|}
\hline
\textbf{Author}&
\centering \textbf{Methodology}& 
\centering \textbf{Dataset used} &
\centering\arraybackslash \textbf{Dice coeff}\\ \hline

Devnathan et al. \cite{9505562} &
\centering Fuzzy C-Means Clustering &
\centering\arraybackslash ISIC-2018 Skin Lesion & 
0.75\\ \hline

Beddad et al. \cite{8966821} &
\centering Fuzzy C-means Clustering &
\centering\arraybackslash Brain MRI
& -\\ \hline

Ronneberger et al. \cite{ronneberger2015u}&
\centering U-Net &
\centering\arraybackslash Electron Microscopic Stacks 
& - \\ \hline

Zhou et al. \cite{8932614} &
\centering U-Net++ &
\centering\arraybackslash Cell Nuclei and Colon Polyp 
& ~0.8 \\ \hline

Jha et al. \cite{8959021} &
\centering ResUNet++ &
\centering\arraybackslash Kvasir-SEG and CVC-ClinicDB 
& - \\ \hline

Yin et al. \cite{9118916} &
\centering deep guidance network &
\centering\arraybackslash - 
& - \\ \hline

Feng et al. \cite{9776250} &
\centering Transformer + UNet & 
\centering\arraybackslash ATLAS 
& 0.82\\ \hline

Huang et al. \cite{9750648} &
\centering TransDE &
\centering\arraybackslash CVC-ClinicDB 
& - \\ \hline

\textbf{Our Work} &
\centering \textbf{LeViT UNet++} & 
\centering\arraybackslash \textbf{UW-Madison GI Tract Image } 
& 0.79\\ \hline

\end{tabular}
}
\end{center}
\label{related}
\end{table}

\section{Methodology}
\begin{figure}[ht]
    \centering
    \includegraphics[width = \linewidth]{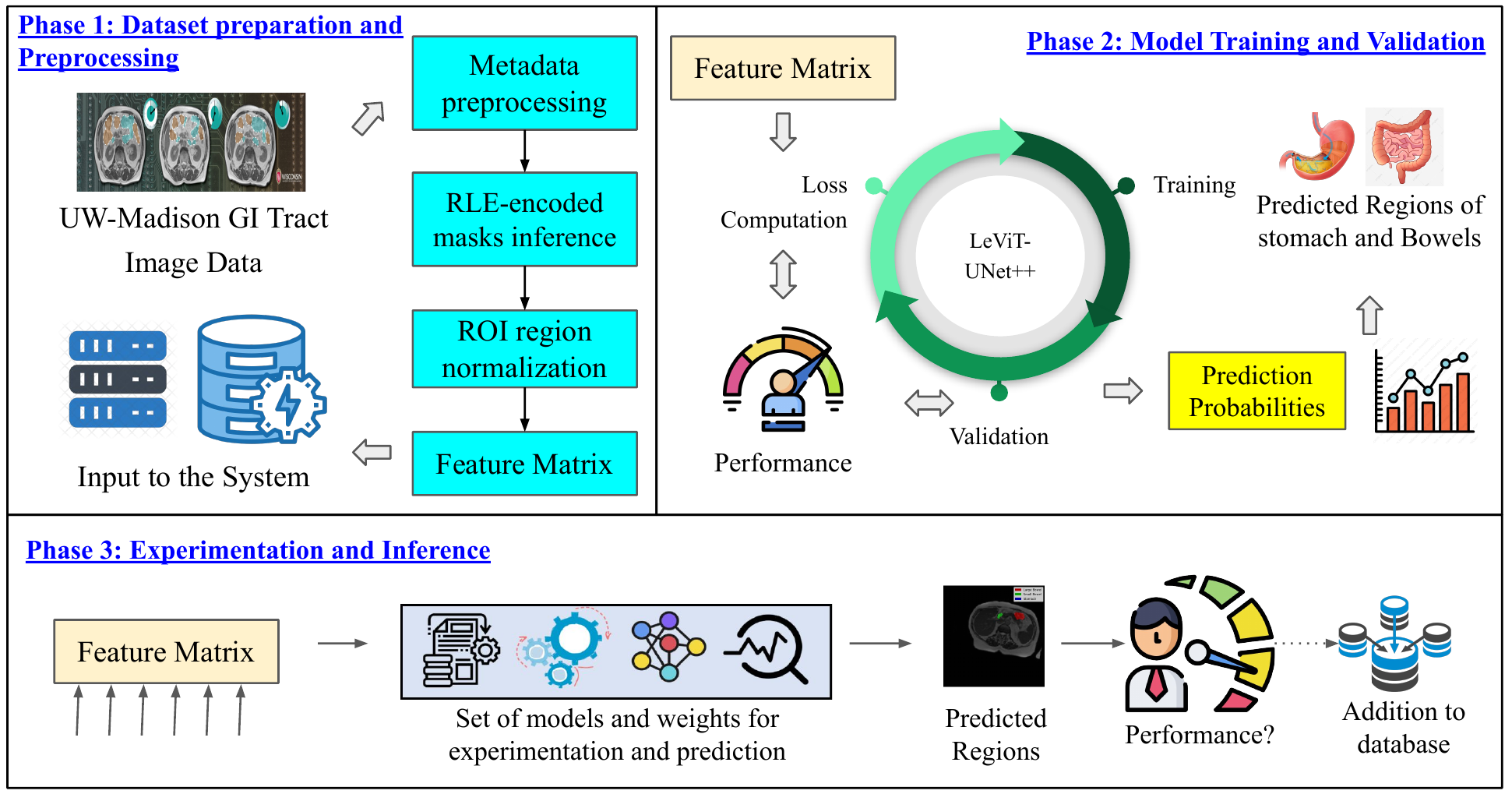}
    \caption{Methodology of the proposed solution}
    \label{fig:methodology}
\end{figure}

We employ a three-phase methodology in our proposed solution. The first phase consists of image and metadata preprocessing. In the subsequent phase, we emphasize the architecture and the hyperparameters used to train the model. In the final phase, we perform exclusive experimentation involving the usage of different architectures for encoders and decoders to justify the effectiveness of our model. The overview of our methodology is depicted in Fig. \ref{fig:methodology}.

\subsection{Image and Metadata Preprocessing}
The images present in the 16-bit grayscale format need to be preprocessed for effective feature extraction. So we employ the process of global contrast normalization for the same. The significance of image normalization is that it helps in faster convergence while training the CNN. We perform the Global Contrast Normalization (GCN) \cite{cheng2014global} to reduce the luminance difference between bright and dark pixels in a given image, as depicted in Algorithm 1. 

\begin{algorithm}
 \caption{Global Contrast Normalization}
 \begin{algorithmic}
 \renewcommand{\algorithmicrequire}{\textbf{Input:}}
 \renewcommand{\algorithmicensure}{\textbf{Output:}}
 \REQUIRE $rawimg, \lambda, \epsilon$\\
 \ENSURE  GCNImage \\ 
  \IF {($Contours \ne None$)}
  \STATE  {\it img = cv2.imread($rawimg$)}
  \STATE  {\it x = np.array(img)} \\
  \STATE  {\it $Average_{x}$ = np.mean(x)} \\ 
  \STATE  {\it Contrast = $\sqrt{(\lambda + np.mean(x^{2})}$} \\
  \STATE  {\it GCN = $s \times x/max(contrast, \epsilon)$} \\ 
  \ENDIF
  \RETURN GCN
 \end{algorithmic} 
 \end{algorithm}
 
\begin{figure*}[t]
    \centering
    \includegraphics[width = \linewidth]{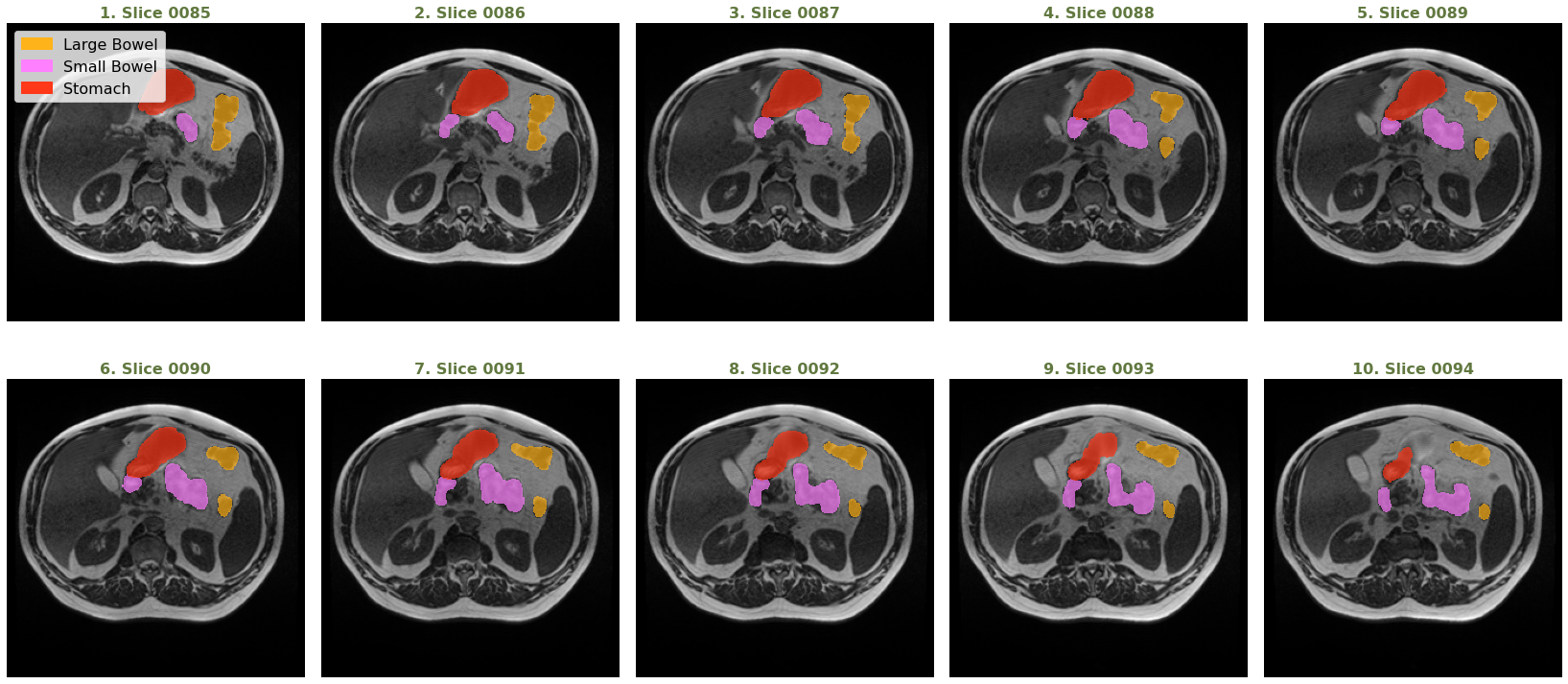}
    \caption{Images and Masks of Case 123, Day 20, Slices 0085-0094}
    \label{fig:mask}
\end{figure*}

 After the image preprocessing step, the subsequent step is embedding the mask from the metadata. A case id has three possibilities: No mask, presence of all organ masks, and presence of only some organ masks. A list of values with various pixel locations and lengths serves as the segmentation in the metadata (where the value is not nan). Fig. \ref{fig:mask} shows the images with the corresponding masks plotted of Case 123, Day 20, and slices 0085-0094. From the above image, we can infer that the segmentation of the stomach and large bowel decreases with each slice, while the segmentation of the small bowel increases with each slice.

\subsection{Model Architecture}
\textbf{LeViT as Encoder:}
We employ the LeViT architecture stated in \cite{graham2021levit} \cite{xu2021levit} as the encoder in this work. This encoder consists of the Transformer Blocks and the Convolutional Blocks. The convolutional blocks consist of four convolutional layers of size 3X3 for dimensionality reduction. The transformer block will receive these feature maps, reducing the amount of known extensive floating-point operations (FLOPs) in transformer blocks.

\textbf{UNet++ as Decoder:} U-Net++ \cite{zhou2018unet++} is considered an enhanced version of U-Net. U-Net consists of an encoder and a decoder shaped in the form of a U that is widely used to train upon fewer samples of data. Primarily, U-Net architecture is a 23-layered architecture consisting of reiterated unpadded convolutions and ReLU and max pooling operations for downsampling. It consists of components like an encoder network, skip connections, bridge, and decoder network. In the case of UNet++, the semantic void between the encoder and decoder networks is filled by the presence of convolutional layers on the skip connections. The gradient flow's enhancement is also seen with the presence of rampant skip connections on the skip pathways. The output from the previous convolution layer of the same dense block is combined with the matching up-sampled output of the lower dense block in the concatenation layer that comes before each convolution layer. This can be illustrated by Eq. \ref{UNet++} where $y^{i,j}$ denotes the feature maps while  H(·) is a convolution operation and U(.) is an activation function:

\begin{equation}
\label{UNet++}
    y^{i,j} = 
    \begin{cases}
       \textit{$H(y^{i-1,j})$}, j=0 \\
       \textit{$H([y^{i,k}], U(u^{i+1,j−1}))$}, j>0
    \end{cases}
\end{equation}

In the subsequent part of the architecture, we emphasize the hyper-parameters used to train the model. The training batch size is 64, while the validation batch size is 128. The initial learning rate is 2e-3 with CosineAnnealingLR as a scheduling technique, and the minimum learning rate is 1e-6. For the model evaluation, we use two metrics: Dice Coefficient and Jaccard Coefficient \cite{setiawan2020image}.

\section{Results and Experimentation}
In this section, we emphasize the dataset's details and exclusive experimentation with different combinations of both the encoder and decoder architectures. This section also discusses the significance of the learning rate's variation concerning each fold in cross-validation.

\subsection{Dataset Used}

\begin{figure}[htbp]
    \centering
    \includegraphics[width = \linewidth]{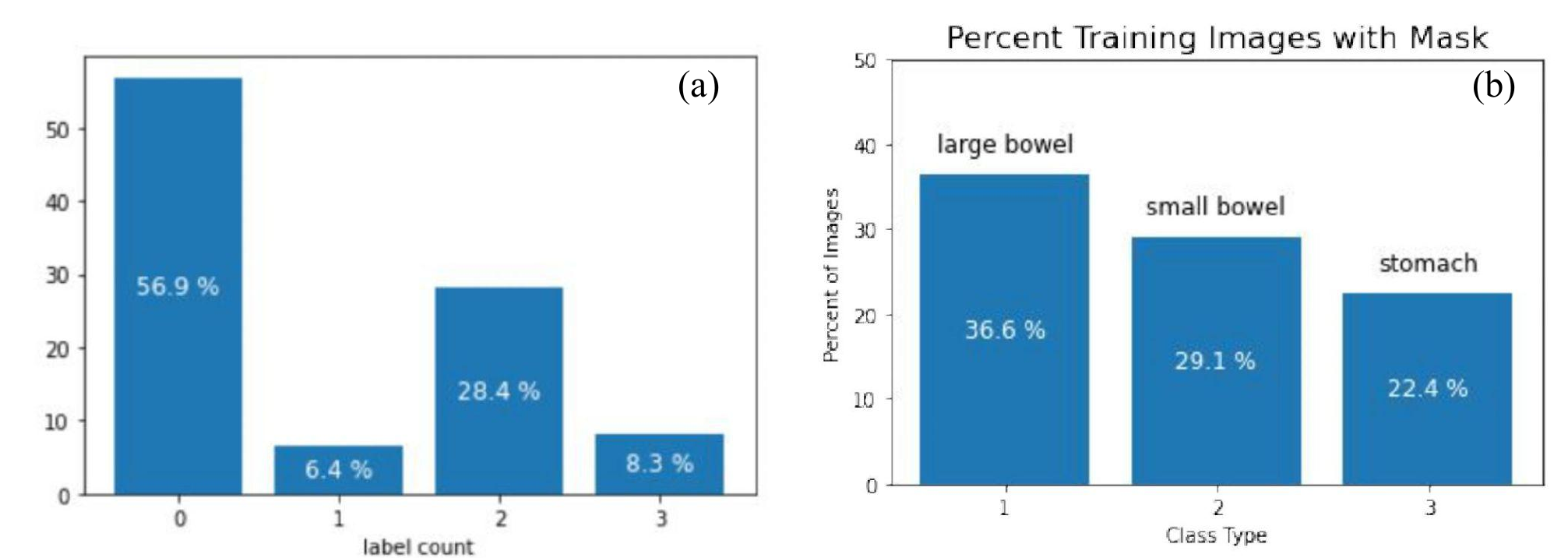}
    \caption{(a) Distribution of Organ Instances in the dataset, (b) Distribution of the segmentation presence in the dataset images}
    \label{fig:EDA}
\end{figure}

We use the UW-Madison GI Tract Image Segmentation dataset to segment the different organs in this work. This dataset consists of about 38000 16-bit grayscale images in the .png format and their annotations in the form of RLE-encoded masks. Multiple sets of scan slices represent each instance (each set is identified by the day the scan took place). Some cases are divided into time segments, while other instances are divided into case segments, with the entire case being in the train set. Fig. \ref{fig:EDA} (a) shows the distribution of the number of organs present in a segment (0,1,2,3), while Fig. \ref{fig:EDA} (b) shows the distribution of each organ in the training set having masks.

\begin{figure*}[htbp]
    \centering
    \includegraphics[width = \linewidth]{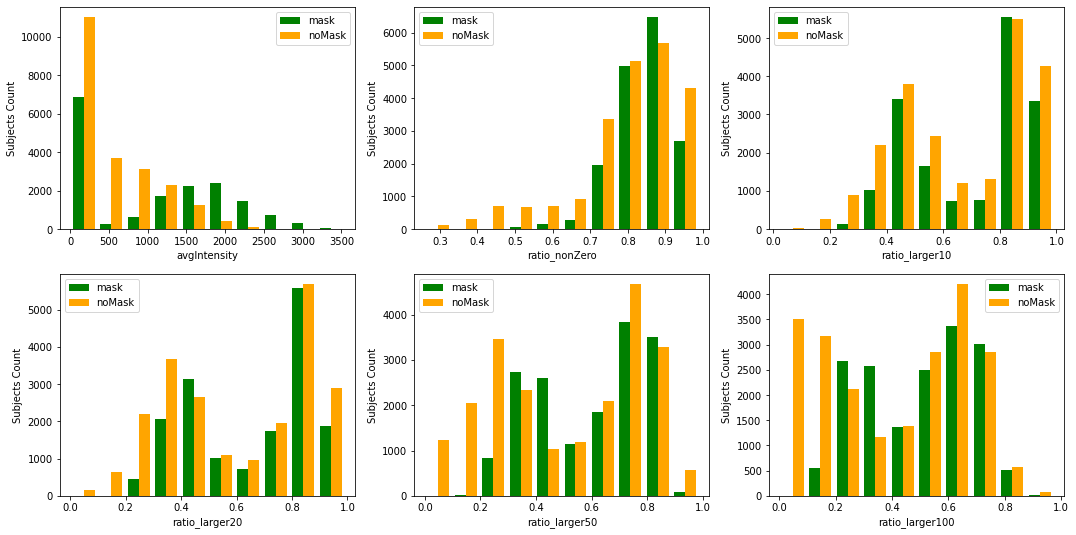}
    \caption{(a) Average Intensity, (b) Ratio of number of non zero pixels to total number of pixels for each image, (c) Ratio of number of pixels that has value larger than 10 to total number of pixels (d) Larger than 20, (e) Larger than 50, (f) Larger than 100 with respect to the subjects count}
    \label{fig:intensity}
\end{figure*}

\begin{figure}[htbp]
    \centering
    \includegraphics[width = \linewidth]{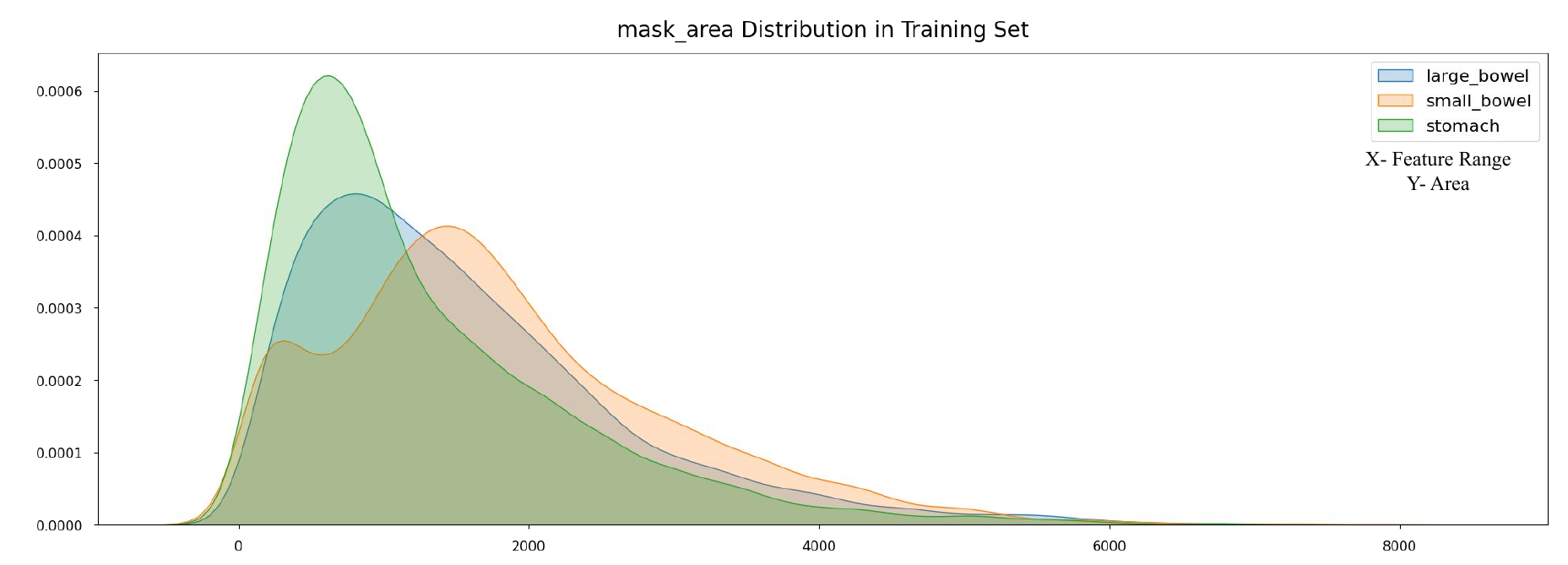}
    \caption{Mask Area Distribution of the dataset}
    \label{fig:area}
\end{figure}

However, the distributions of mask areas among classes are very diverse. The areas that depict the small bowel are comparatively more extensive than those that depict the Large Bowel. It can be inferred that the Small Bowel consists of more area in the X-Y dimension while the large bowel is distributed over the Z dimension. Also, the distribution of the mask area for the stomach class and the number of annotations indicate that it is the smallest class. Fig. \ref{fig:area} indicates the Mask Area Distribution of the dataset.

Also, we analyze the intensity distribution of the images present in the dataset by defining metrics that correspond to the average intensity of an image. The significance of this process is that images that depict the initial stages of treatment can be easily identified. However, we must establish a safety buffer because the provided set is unlikely to be entirely representative. Also, we define other metrics like \textit{ratio-nonZero}, which is the proportion of non-zero pixels to each image's pixels. To thoroughly analyze the dataset from an intensity perspective, we also illustrate ratio-larger10 as the proportion of the number of pixels with a value larger than 10, to the total number of pixels, for each image. Similarly, we also depict other metrics like \textit{ratio-larger20, ratio-larger50, ratio-larger100} in Fig. \ref{fig:intensity}.

\begingroup
\setlength{\linewidth}{8pt} 
\begin{table*}[t]
	\centering
	\caption{Performance comparison of different architecture combinations for medical image segmentation}
	\label{arctable}
	\begin{tabular}{|m{4em}|m{4em}|m{4em}|m{4em}|m{4em}|m{4em}|m{4em}|m{4em}|m{4em}|m{4em}|m{4em}|m{4em}|m{4em}|}
		\hline
		\multicolumn{1}{|c}{\textbf{}} & \multicolumn{4}{|c|}{\textbf{LeViT-128s}}  & \multicolumn{4}{c|}{\textbf{LeViT-192}}  & \multicolumn{4}{c|}{\textbf{LeViT-384}} \\
		\hline

		\centering \textbf{CNN} &
		\centering {Train Loss} &
		\centering Valid Loss &
		\centering Valid Dice &
		\centering Valid Jaccard &
		\centering {Train Loss} &
		\centering Valid Loss &
		\centering Valid Dice &
		\centering Valid Jaccard &
		\centering {Train Loss} &
		\centering Valid Loss &
		\centering Valid Dice &
		\centering Valid Jaccard \tabularnewline 
		\hline 

		\centering \textbf{U-Net} &
		\centering {0.144} &
		\centering 0.192 &
		\centering 0.703 &
		\centering 0.633 &
		\centering 0.153 &
		\centering 0.200 &
		\centering 0.698 &
		\centering 0.629 &
		\centering 0.124 &
		\centering 0.189 &
		\centering 0.723 &
		\centering 0.714 \tabularnewline 
		\hline 

		\centering \textbf{U-Net++} &
		\centering {0.137} &
		\centering 0.188 &
		\centering 0.711 &
		\centering 0.641 &
		\centering 0.140 &
		\centering 0.187 &
		\centering 0.705 &
		\centering 0.628 &
		\centering 0.089 &
		\centering 0.134 &
		\centering 0.795 &
		\centering 0.728 \tabularnewline 
		\hline

	\end{tabular}
\end{table*}
\endgroup

\subsection{Performance Metrics of LeViT384-UNet++}
\begin{figure}[htbp]
    \centering
    \includegraphics[width = \linewidth]{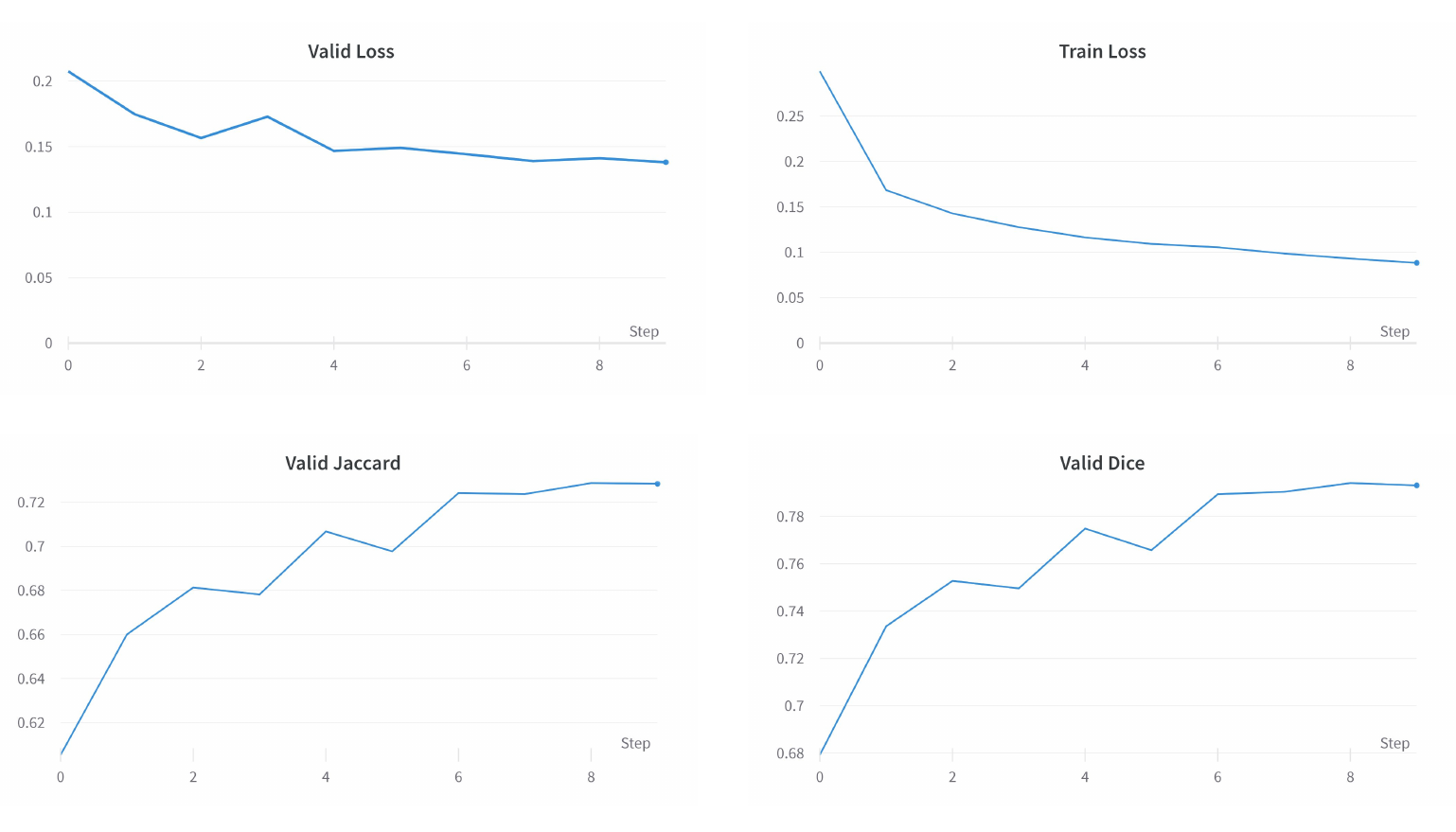}
    \caption{(a) Validation Loss, (b) Training Loss, (c) Validation Dice Coefficient, (d) Validation Jaccard Coefficient for each training epoch}
    \label{fig:base}
\end{figure}

Once trained on our input data, it is deemed essential to evaluate our model. A methodological error is incorporated if the model retains the parameters of a periodic function and is experimented on the same data. The prediction scores would remain perfect on known labels, and the model's performance would still be unsatisfactory on unseen data. This condition is also known as overfitting. So to prevent overfitting, it is always deemed essential to split out a chunk of data into the test/validation set. However, there is a probability of overfitting the test/validation set due to tweaking the existing parameters until the estimator performs correctly. So to address this situation, we perform K-Fold Stratified Cross-Validation \cite{wong2017dependency}. The objective of the K-Fold Stratified Cross-Validation is that the data is split into K folds. Training is performed on K-1 folds while testing/validation is performed on the remaining fold, resulting in a higher performance of the model. Also, we evaluate the Validation Dice Coefficient and Jaccard Coefficient for each fold as depicted in Table \ref{base}. As depicted in Table \ref{base}, the highest Validation Dice Coefficient and Jaccard Coefficient were found to be 0.79543 and 0.72821, respectively. Also, the training loss and validation loss were recorded to be 0.08934 and 0.13474, respectively. Fig. \ref{fig:base} shows the variation of the above metrics for each training epoch.

\begin{table}[htbp]
\caption{Performance Metrics of LeViT384-UNet++}
\begin{center}
\resizebox{\columnwidth}{!}{%
\begin{tabular}{|c|c|c|c|c|c|}
\hline
\textbf{Fold}&
\textbf{Train Loss}& 
\textbf{Valid Loss}&
\textbf{Valid Dice} &
\textbf{Valid Jaccard}&
\textbf{LR}\\ \hline

1&
0.08652&
0.13891&
0.79105&
0.72793&
0.00098 \\ \hline

2&
0.08934&
0.13474&
0.79543&
0.72821&
0.00094 \\ \hline

3&
0.08843&
0.13811&
0.79313&
0.72846&
0.00099 \\ \hline

4&
0.10058&
0.1421&
0.78583&
0.71505&
0.00111 \\ \hline

\end{tabular}
}
\end{center}
\label{base}
\end{table}
\vspace{-10px}

\subsection{Variation of Learning Rate}
\begin{figure}[htbp]
    \centering
    \includegraphics[width = \linewidth]{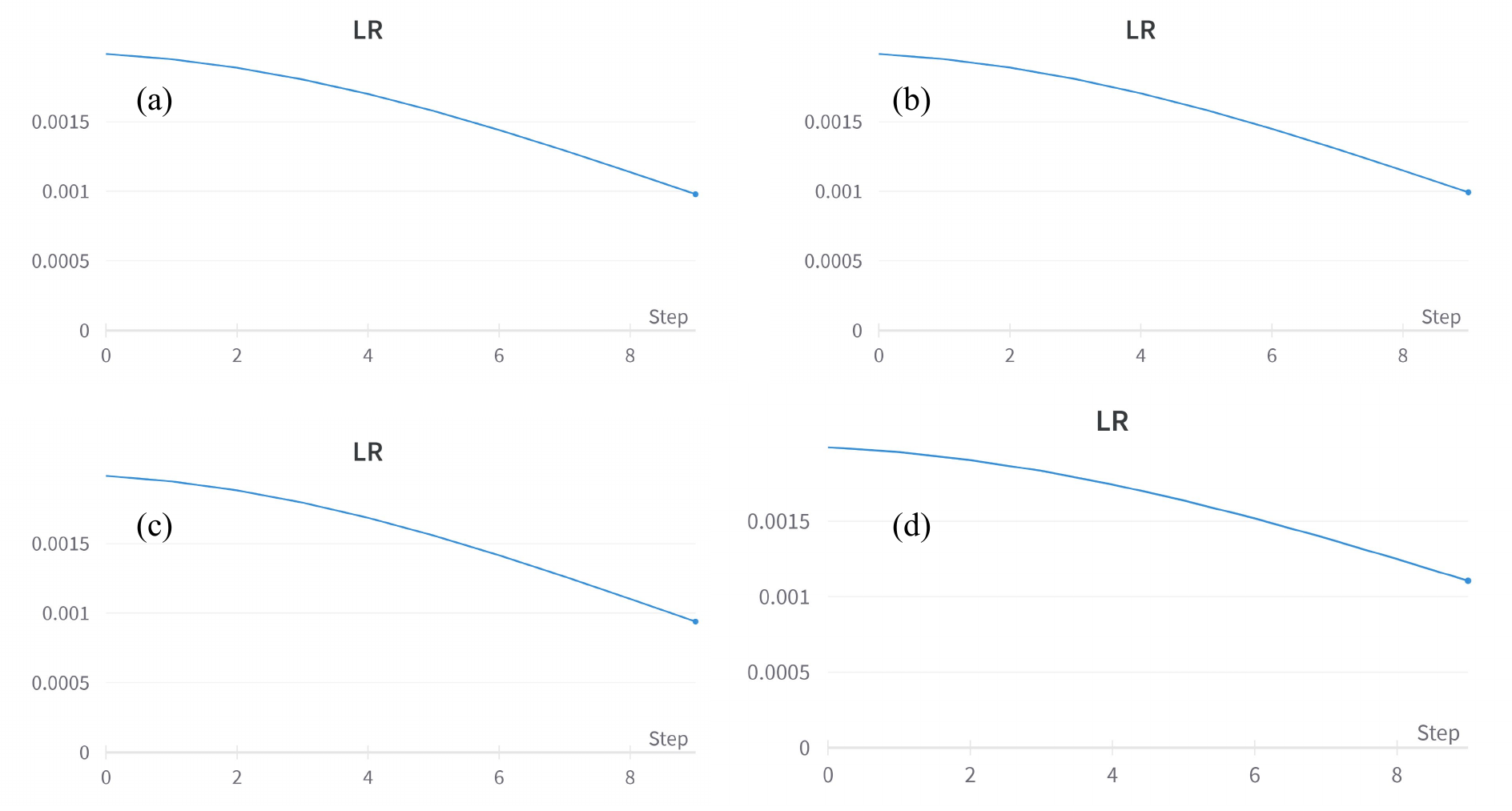}
    \caption{Variation of Learning Rate for each fold}
    \label{fig:lr}
\end{figure}

The majority of researchers concur that neural network models are challenging to train. This can be inferred from the fact that there are a large number of hyperparameters that need to be specified and optimized. One example of such a hyperparameter is the learning rate. Since the learning rate defines how effectively the model fits the data, finding an appropriate learning rate with each epoch is essential. So to perform this task, we use the {\it CosineAnnealingLR} scheduling technique. In this technique, the learning rate begins with a high value, and subsequently, it rapidly reduces the learning rate to a number close to 0 and then raises the learning rate again. Fig. \ref{fig:lr} depicts the learning rate variation for each epoch for each fold. 

\subsection{Performance comparison of different architecture combinations}
\begin{figure}[htbp]
    \centering
    \includegraphics[width = \linewidth]{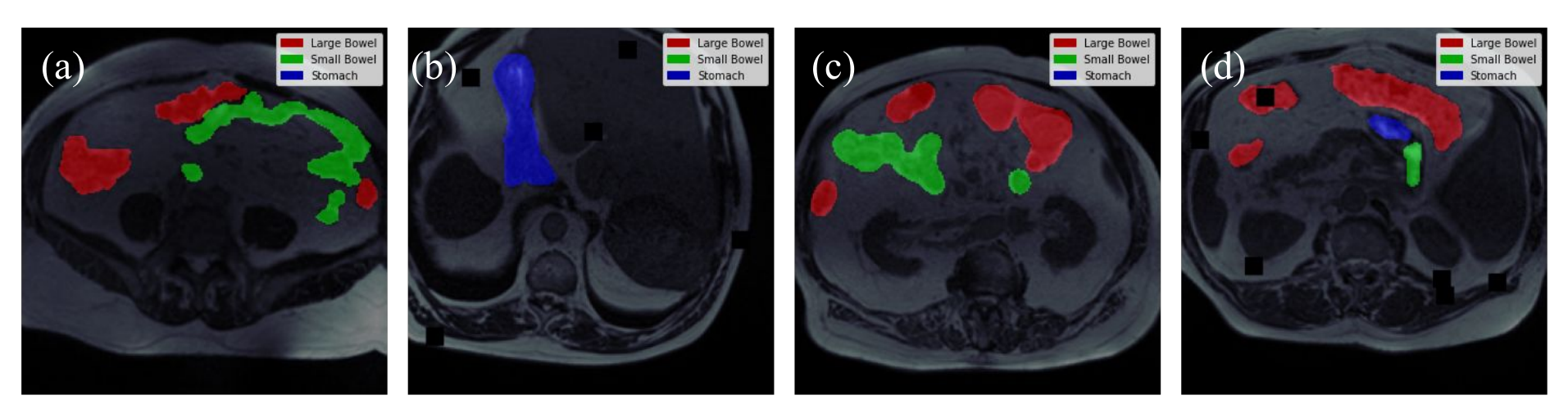}
    \caption{(a) Large and Small Bowels, (b) Stomach, (c) Large and Small Bowels, (d) All Organs}
    \label{fig:my_label}
\end{figure}

In this section, we experiment with the different combinations of encoders and decoders, comparing the different metrics in each scenario. In the case of the decoder, we use the different versions of Le-ViT, which include LeViT-UNet-128s, LeViT-UNet-192, and LeViT-UNet-384. Subsequently, in the case of the encoder, we experiment with two architectures: U-Net and U-Net++. The best combination is LeViT-UNet-384 and U-Net++, as depicted in Table \ref{arctable}. This section's final component depicts our architecture's performance on unseen data. To prove the efficiency of our model, we have equal samples containing all or a single mask in an image. Fig. \ref{fig:my_label} (a) and (c) show samples of prediction on images having clear large and bowel segments, while Fig. \ref{fig:my_label} (b) shows the predicted segment of the stomach. Finally, Fig. \ref{fig:my_label} (d) clearly shows the predictions on images with all the organs segmented. 

\section{Conclusion}
Given the factors of extensive image data and automated segmentation requirements, there is a need for scalable, robust, and rapid automated deep-learning-based medical image segmentation. This paper addresses the problem from a computational efficiency perspective. We have performed the segmentation of the different organs from an MRI scan using a hybrid CNN-Transformer-based architecture. The solution is meticulously evaluated on many examples with no or partial organ segments and systematically studying the importance of different features. To improve the model's performance, the architecture can be further modified and improved for unseen domains as well. Also, further developments include the investigation of security issues relating to the
machine-learning model and the image database.

\end{document}